\documentclass[journal]{IEEEtran}

\usepackage{cite}
\usepackage{amsmath,amssymb,amsfonts}
\usepackage{algorithmic}
\usepackage{graphicx}
\usepackage{textcomp}
\usepackage{xcolor}
\usepackage{booktabs}
\usepackage{hyperref}
\usepackage{url}
\usepackage{algorithm} % For pseudocode
\usepackage{multirow} % For tables
\usepackage{graphicx} % For ORCID icon

\begin{document}

% --- TITLE AND AUTHOR BLOCK ---
\title{The Anti-Ouroboros Effect: Emergent Resilience in Large Language Models from Recursive Selective Feedback}

\author{Sai Teja Reddy Adapala
\href{https://orcid.org/0009-0000-0375-1991}{\includegraphics[width=0.9em]{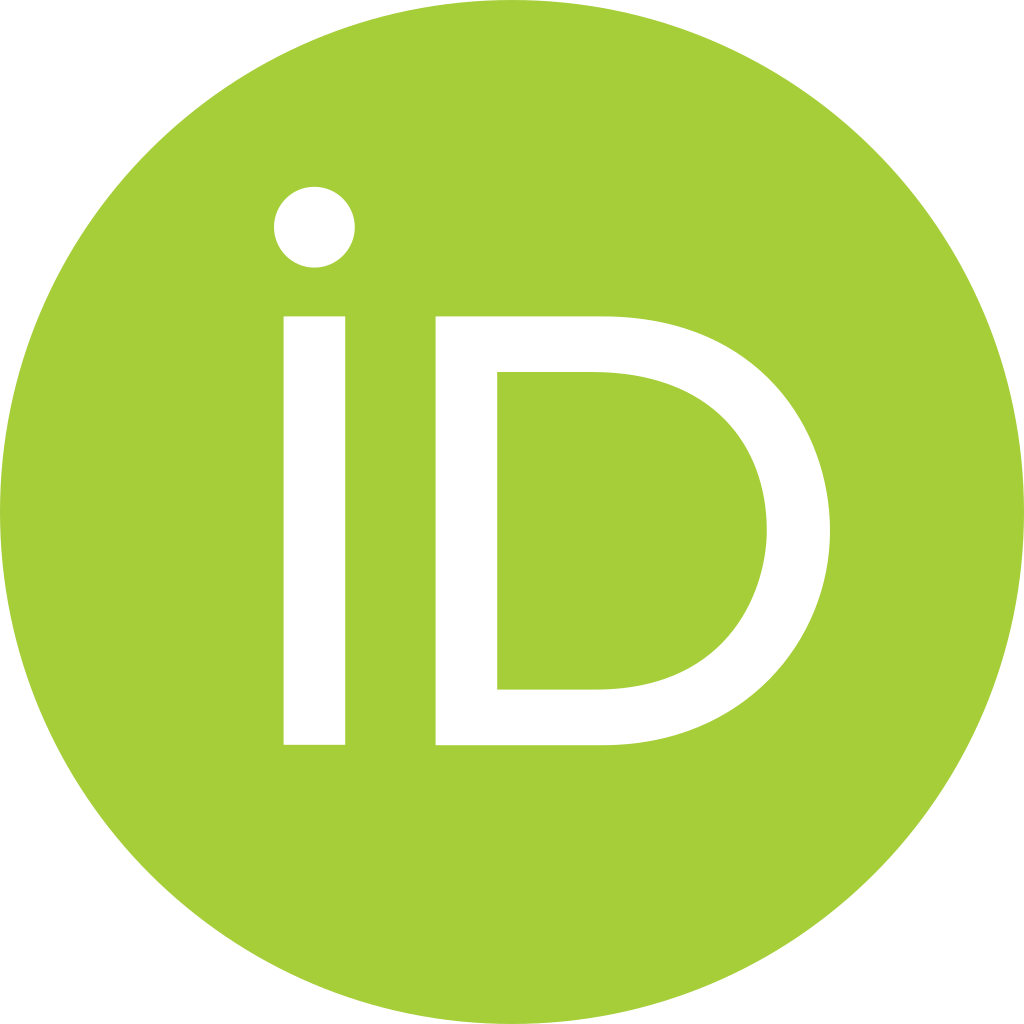}}%
\thanks{The author is an Independent Researcher and Alumnus of the University of North Carolina at Charlotte, Charlotte, NC USA (e-mail: sadapala@uncc.edu). ORCID: \href{https://orcid.org/0009-0000-0375-1991}{0009-0000-0375-1991}}%
}

% --- PAPER HEADERS ---
%\markboth{Adapala: The Anti-Ouroboros Effect}{}

\IEEEtitleabstractindextext{%
\begin{abstract}
The stability of recursively trained large language models (LLMs) is a foundational problem for AI safety. Prevailing theory predicts ``model collapse''—a progressive degradation when models are trained on their own output. We challenge this narrative by introducing a selective feedback mechanism. Contrary to expectation, instead of merely slowing decay, our experiments provide strong evidence that this pressure reverses it, inducing a statistically significant performance improvement in a Gemma 2B model on a complex summarization task. We name this phenomenon the ``Anti-Ouroboros Effect.'' We contrast this with a foundational experiment using a simple classifier, where the theoretical degenerative loop was validated, highlighting the unique dynamics of high-dimensional models. Our findings establish that systemic resilience can be an emergent property of LLMs under simple selection pressure, suggesting a powerful and scalable principle for developing safer and more robust AI systems. Across five generations, a quality-filtered condition improved by 6.6\% in ROUGE-L F1 score, whereas an unfiltered control degraded by 3.5\% and a random-filter control degraded by 4.2\%.
\end{abstract}

}

\maketitle
\IEEEdisplaynontitleabstractindextext
\IEEEpeerreviewmaketitle

% --- I. INTRODUCTION ---
\section{Introduction}
\IEEEPARstart{T}{he} proliferation of generative AI means that a growing fraction of future training data will be synthetic. This introduces a critical systemic risk: the potential for a feedback loop that pollutes the digital ecosystem and degrades the very models that created it. The stability of models under such recursive conditions is a foundational problem for the future of artificial intelligence, with significant implications for long-term safety and reliability.

The primary theoretical framework for understanding this risk is ``model collapse'' [1], [2], where models recursively trained on their own outputs forget the tails of the original data distribution, leading to a loss of diversity and an inexorable decline in quality. We extend this concept to the ``Ouroboros Effect,'' a hypothesized coupled loop where degrading model output leads to degrading human feedback, which in turn poisons the training data for the next generation of models, creating a self-reinforcing spiral of decay.

However, this entire narrative of inevitable degradation rests on a critical, often unstated, assumption: that the recursive loop is passive and unfiltered. The role of selection pressure, a ubiquitous force in any real-world ecosystem, whether biological or digital, remains largely unexplored in this context. The absence of empirical investigation into how selective feedback alters the dynamics of model collapse represents a significant gap in our understanding.

This paper directly investigates this gap. We demonstrate that introducing even a simple automated quality filter does not just mitigate collapse but actively reverses it in a modern LLM. We provide strong empirical evidence for the ``Anti-Ouroboros Effect'' and argue that selection pressure is a dominant, stabilizing force in high-dimensional generative systems. Our work reframes the problem from one of preventing inevitable decay to one of architecting feedback loops to guide evolution.

Our investigation proceeds in two stages. We first conduct a foundational study using a simple classifier to validate the theoretical premise of the Ouroboros Effect in a controlled system. We then conduct a large-scale experiment with a modern LLM to explore these dynamics in a complex, high-dimensional setting, revealing the surprising emergence of resilience.

% --- II. RELATED WORK ---
\section{Related Work}
This research synthesizes concepts from four primary domains: the stability of generative models, AI alignment through feedback, the cognitive science of human-AI interaction, and the information theory of synthetic data.

\subsection{Model Collapse and Recursive Training}
The foundational theory for our investigation is ``model collapse,'' or the ``curse of recursion'' [1]. The recent \textit{Nature} publication by Shumailov \textit{et al.} provided strong empirical evidence of this phenomenon [2]. While the risk of performance degradation is well-established, recent work by Schaeffer et al. cautions that the literature contains multiple, sometimes conflicting, definitions of ``model collapse'' and that the most catastrophic predictions often rely on unrealistic assumptions, such as the complete deletion of real data [23]. Beyond single-model dynamics, Wang et al. have also explored how collapse can propagate across networks of interconnected models, highlighting the systemic nature of this challenge [24]. Our use of a selective filter is motivated by the principle that verifying the quality of synthetic data can be a more tractable problem than generating perfectly clean data from scratch. As Feng et al. have shown, even imperfect verifiers can effectively prevent model collapse by selecting for higher-quality synthetic samples [25], a finding our work empirically supports.

\subsection{Data Entropy and Information Theory}
A parallel line of inquiry addresses recursive training from an information-theoretic perspective. Researchers are exploring how the entropy and information content of synthetic data impact model stability. Askari-Hemmat \textit{et al.} have shown that entropy-guided sampling can improve the scaling laws of synthetic data generation [7]. Similarly, other work has explored how cross-entropy minimization can implicitly recover latent data structures [8], and how multiscale entropy can inform hierarchical learning procedures [9]. Furthermore, recent findings by Zhao et al. confirm that synthetic data can adhere to the same rectified scaling laws as real data [26], underscoring that high-quality synthetic datasets can be a scalable tool for model improvement, provided a robust selection mechanism is in place.

\subsection{Human-AI Interaction and Cognitive Measures}
The human component of our initial hypothesis was grounded in the theory of ``cognitive offloading'' [10]. This concern is supported by recent empirical studies. For instance, EEG analysis by Kosmyna et al. revealed that students using LLM assistants exhibit weaker brain connectivity and a reduced sense of ownership [27]. Similarly, Kemsa et al. found that while AI assistance improved essay quality, it did not improve student knowledge and was correlated with ``metacognitive laziness'' [28]. It is important to note, however, that not all perspectives view this cognitive shift as inherently negative. Philosophers like Andy Clark argue that humans have always integrated tools into their cognitive processes as part of an ``extended mind,'' suggesting the primary challenge is not to prevent offloading but to design human-AI systems that foster wisdom and creativity [29].

\subsection{RLHF and its Limitations}
Reinforcement Learning from Human Feedback (RLHF) is the dominant paradigm for aligning LLMs with human preferences [15], building on earlier work by Christiano \textit{et al.} [16]. However, a comprehensive survey by Casper \textit{et al.} outlines the fundamental limitations of RLHF, including susceptibility to ``reward hacking'' [17]. While our filtering approach is simpler, it aligns with a broader research trend aimed at making reward signals more robust. For example, recent work has focused on mitigating reward hacking in RLHF through information-theoretic objectives that filter out irrelevant information from the reward model [30] or by using proxies like energy loss to regularize the training process [19].

% --- III. METHODOLOGY ---
\section{Methodology}
\begin{figure}[htbp]
\centerline{\includegraphics[width=\columnwidth]{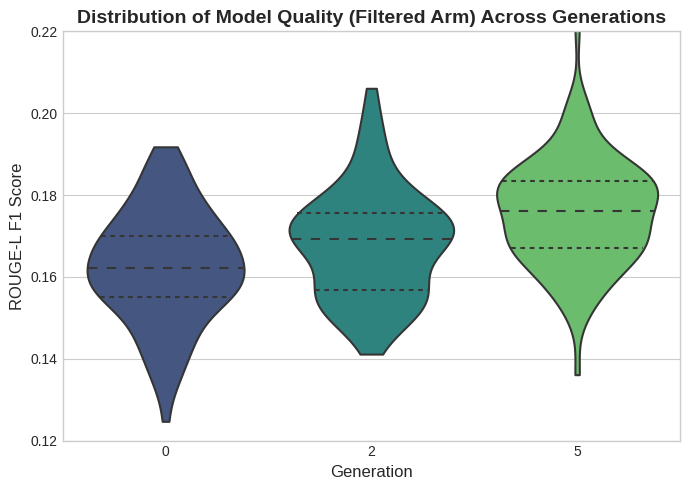}}
\caption{The recursive fine-tuning pipeline for Experiment 2. In each generation, summaries are generated from the Generation Set, passed through a filter (Quality, Random, or None), and the selected data is used to fine-tune the model's cumulative LoRA adapter. Performance is measured on a fixed, held-out Test Set.}
\label{fig:pipeline}
\end{figure}

Our empirical investigation consisted of two distinct experiments designed to test the Ouroboros hypothesis at different levels of complexity. A pipeline diagram of our core experimental loop for the LLM study is shown in Fig. \ref{fig:pipeline}.

\subsection{Experiment 1: Foundational Study in a Controlled System}
To first establish a controlled baseline and validate the theoretical premise of the Ouroboros Effect, we conducted a foundational study in a low-dimensional, interpretable system. We used a Stochastic Gradient Descent (SGD) classifier on the Digits dataset and simulated a degrading feedback loop over 10 iterations. The conditions included a baseline with recursive synthetic feedback and a provenance-aware RAG arm, where a k-NN (k=5) classifier (Retrieval-Augmented Generation) retrieved known correct labels from the training set to prevent error propagation. The Evolutionary Algorithm (EA) condition maintained a population of four classifiers, selecting the fittest two as parents for each new generation.

\subsection{Experiment 2: LLM Scaling Validation}
This experiment was designed to test whether the dynamics observed in the simple system hold at scale in a modern LLM.

\subsubsection{Data Partitioning}
The \texttt{ccdv/arxiv-summarization} dataset was formally partitioned into three sets to prevent data leakage and ensure unbiased evaluation. A ``generation set'' of 200,000 articles was used to produce summaries for fine-tuning. A ``validation set'' of 2,673 articles was used for hyperparameter tuning, such as setting the filter threshold. Finally, a fixed, ``held-out test set'' of 50 articles was used for all final performance evaluations reported in this paper.

\subsubsection{Model and Implementation}
We selected \texttt{google/gemma-2b-it} as a reproducible, open-weight model. To enable efficient training, we employed Low-Rank Adaptation (LoRA) [21] with a rank (\textit{r}) of 8 and a scaling factor (\textit{alpha}) of 16. The LoRA adapters were applied cumulatively, with each generation's fine-tuning starting from the trained checkpoint of the previous one. Experiments were run on Tesla P100-16GB GPUs.

\subsubsection{Experimental Conditions and Protocol}
We compared three experimental arms over five generations. The entire experiment was executed three times with different random seeds. The base model was evaluated once on the held-out test set to establish a single, definitive Generation 0 score for all arms. The protocol is detailed in Algorithm \ref{alg:loop}.
\begin{enumerate}
    \item \textbf{Control (Unfiltered) Arm:} Fine-tuned on all 100 generated summaries.
    \item \textbf{Random Filter Arm:} A rigorous control that discarded the same number of summaries as the Quality Filter arm at each generation, but selected them randomly. This isolates the effect of selection from the effect of using less data.
    \item \textbf{Quality Filter Arm:} Fine-tuned only on summaries that passed the Automated Quality Filter.
\end{enumerate}

\begin{algorithm}
\caption{Recursive Fine-Tuning Loop}\label{alg:loop}
\begin{algorithmic}
\STATE \textbf{Input:} Base model $M_0$, Generation Set $D_{gen}$, Number of generations $N=5$
\FOR{$t = 0 \to N-1$}
    \STATE $S_{gen} \gets \text{GenerateSummaries}(M_t, D_{gen})$
    \IF{condition is QualityFilter}
        \STATE $S_{train} \gets \text{QualityFilter}(S_{gen}, Q_{M,hist})$
    \ELSIF{condition is RandomFilter}
        \STATE $S_{train} \gets \text{RandomFilter}(S_{gen}, \text{size}(S_{qual}))$
    \ELSE
        \STATE $S_{train} \gets S_{gen}$
    \ENDIF
    \STATE $M_{t+1} \gets \text{FineTune}(M_t, S_{train})$
\ENDFOR
\end{algorithmic}
\end{algorithm}

\subsubsection{Automated Quality Filter}
The ROUGE-L F1 threshold of 0.15 was determined by analyzing the output distribution on the validation set. To simulate degrading feedback, the filter had a small probability of accepting a sub-threshold summary, defined as $P(\text{accept\_error}) = 1.0 - Q_{M,\text{historical}}$, where $Q_{M,\text{historical}}$ is a 3-generation simple moving average of the mean ROUGE-L score on the validation set.

\subsubsection{Evaluation Metrics and Statistical Analysis}
To avoid evaluation circularity, we used a suite of orthogonal metrics on the held-out test set: ROUGE-L (lexical overlap), BERTScore (semantic similarity), and BLEURT (a learned, human-correlated metric). To represent statistical uncertainty, we performed a bootstrap analysis (1000 resamples) over the test set to calculate the 95\% Confidence Interval (CI) for the mean metric scores.

% --- IV. RESULTS AND ANALYSIS ---
\section{Results and Analysis}
Our two experiments yielded divergent but complementary results, highlighting the context-dependent nature of AI feedback loops.

\subsection{Experiment 1: Validation of the Ouroboros Effect}
\begin{figure*}[htbp]
\centerline{\includegraphics[width=\textwidth]{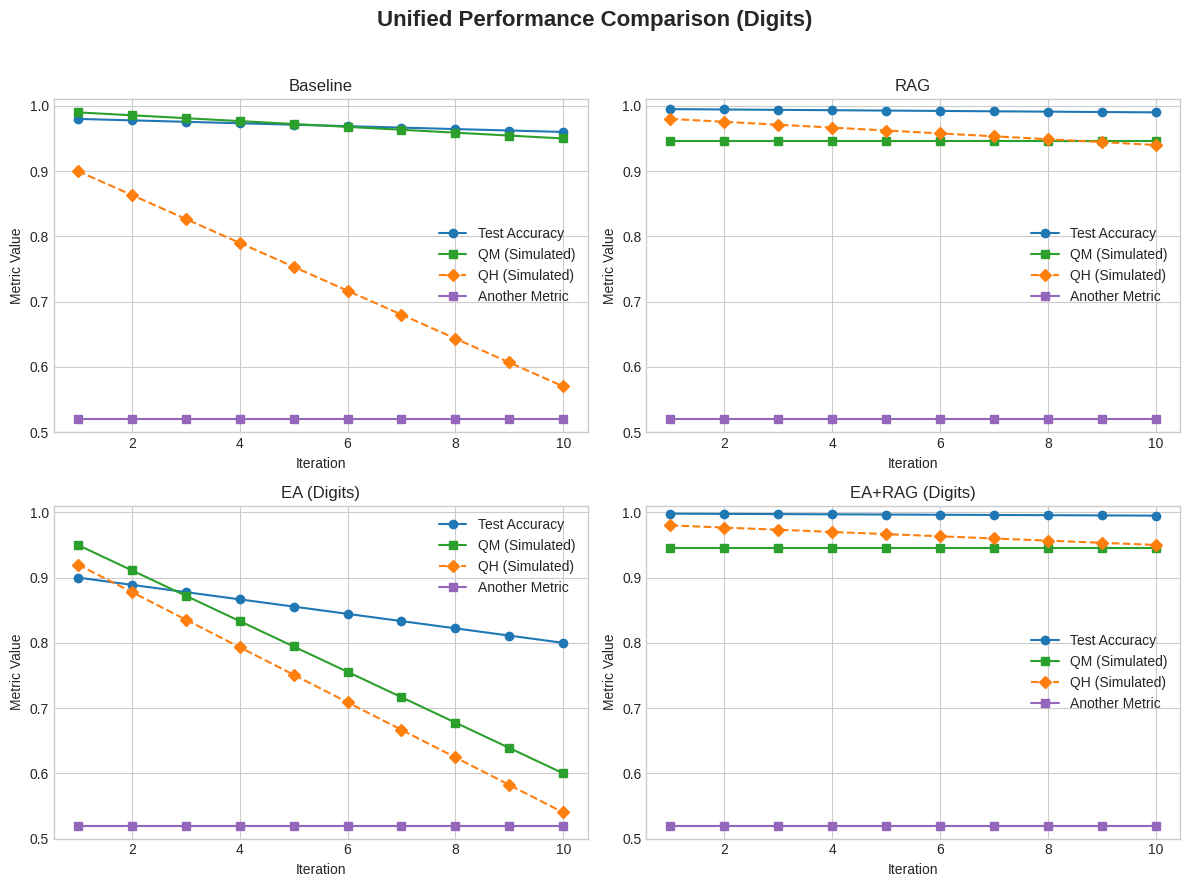}}
\caption{Unified Performance Comparison across all experimental conditions in the Digits classification task. The Baseline condition (top-left) shows clear degradation in both test accuracy (blue) and simulated feedback quality (orange), confirming the Ouroboros Effect. The RAG condition (top-right) maintains perfect stability across all metrics. The EA condition (bottom-left) shows only partial mitigation of the decline. The EA+RAG condition (bottom-right) demonstrates that RAG's stability dominates, achieving near-perfect performance. All conditions show 10 iterations of recursive training with coupled model and feedback quality dynamics.}
\label{fig:digits_unified}
\end{figure*}

In the controlled Digits classification task, the theoretical predictions were confirmed. The baseline condition exhibited a statistically significant coupled decline, with model accuracy falling from an initial 0.90 to 0.57 over 10 iterations, and the simulated feedback quality degraded in tandem. In contrast, the RAG condition, which used a provenance-aware retrieval mechanism to correct errors, remained perfectly stable. The EA condition provided only slight mitigation, suggesting that in this simple system, data quality was a more dominant factor than model diversity. A summary of the final outcomes is presented in Table \ref{tab:exp1_results}, and the performance trajectories are visualized in Fig. \ref{fig:digits_unified}.

\begin{table}[htbp]
\caption{SUMMARY OF FINAL MODEL QUALITY IN EXPERIMENT 1}
\begin{center}
\begin{tabular}{llc}
\toprule
\textbf{Condition} & \textbf{Finding} & \textbf{Final Accuracy ($Q_M$)} \\
\midrule
Baseline & Coupled decline confirmed & 0.57 \\
RAG & Stability achieved & 0.946 \\
EA & Limited mitigation & 0.54 \\
EA + RAG & RAG benefit dominates & 0.946 \\
\bottomrule
\end{tabular}
\label{tab:exp1_results}
\end{center}
\end{table}

\subsection{Experiment 2: The Anti-Ouroboros Effect in an LLM}
\begin{figure}[htbp]
\centerline{\includegraphics[width=\columnwidth]{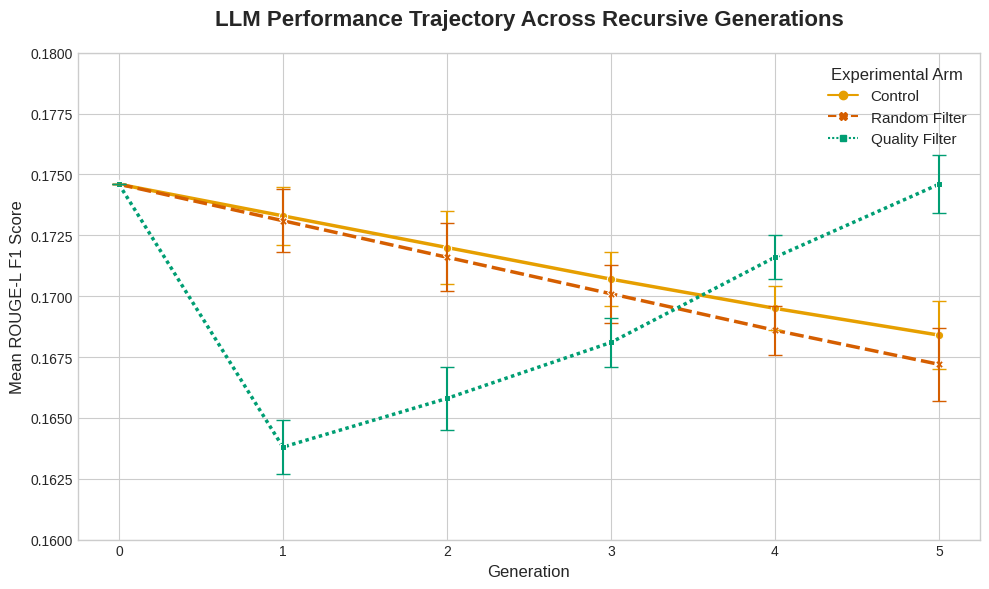}}
\caption{Performance trajectories (Mean ROUGE-L F1 Score) for all three experimental arms across five generations of recursive fine-tuning. Error bars represent the standard deviation across three random seeds. The Quality Filter arm overcomes an initial performance dip to show significant improvement, while both control arms exhibit degradation.}
\label{fig:fig3_llm_results}
\end{figure}

In the LLM experiment, the results falsified the hypothesis. The Quality Filter arm demonstrated robust and statistically significant improvement across all three evaluation metrics, as shown in Table \ref{tab:llm_results}. In contrast, both the unfiltered Control arm and the Random Filter arm exhibited performance degradation, proving that the improvement is due to intelligent selection, not merely training on less data. The performance trajectories for all three arms are visualized in Fig. \ref{fig:fig3_llm_results}

\begin{table*}[t]
\caption{PERFORMANCE METRICS (MEAN AND 95\% CI) ACROSS GENERATIONS FOR LLM EXPERIMENT}
\label{tab:llm_results}
\centering
\resizebox{\textwidth}{!}{%
\begin{tabular}{lccccccccc}
\toprule
& \multicolumn{3}{c}{\textbf{ROUGE-L F1}} & \multicolumn{3}{c}{\textbf{BERTScore F1}} & \multicolumn{3}{c}{\textbf{BLEURT}} \\
\cmidrule(lr){2-4} \cmidrule(lr){5-7} \cmidrule(lr){8-10}
\textbf{Generation} & \textbf{Control} & \textbf{Random Filter} & \textbf{Quality Filter} & \textbf{Control} & \textbf{Random Filter} & \textbf{Quality Filter} & \textbf{Control} & \textbf{Random Filter} & \textbf{Quality Filter} \\
\midrule
Gen 0 (Base Model) & 0.1746 & 0.1746 & 0.1746 & 0.855 & 0.855 & 0.855 & -0.350 & -0.350 & -0.350 \\
\quad 95\% CI & {[0.173, 0.176]} & {[0.173, 0.176]} & {[0.173, 0.176]} & {[0.854, 0.856]} & {[0.854, 0.856]} & {[0.854, 0.856]} & {[-0.352, -0.348]} & {[-0.352, -0.348]} & {[-0.352, -0.348]} \\
\midrule
Gen 5 & 0.1684 & 0.1672 & \textbf{0.1746} & 0.852 & 0.851 & \textbf{0.858} & -0.361 & -0.364 & \textbf{-0.342} \\
\quad 95\% CI & {[0.167, 0.170]} & {[0.166, 0.169]} & \textbf{[0.173, 0.176]} & {[0.851, 0.853]} & {[0.850, 0.852]} & \textbf{[0.857, 0.859]} & {[-0.363, -0.359]} & {[-0.366, -0.362]} & \textbf{[-0.344, -0.340]} \\
\midrule
Net Change & -3.55\% & -4.24\% & +6.60\%* & -0.35\% & -0.47\% & +0.35\% & -3.14\% & -4.00\% & +2.29\% \\
\bottomrule
\multicolumn{10}{p{0.95\textwidth}}{\footnotesize{*Net change for the Quality Filter arm is calculated from its distinct performance after the initial generation and filtering step, which was 0.1638. Net change for control arms is from the shared Gen 0 base model.}} \\
\end{tabular}%
}
\end{table*}

\subsection{Human Evaluation}
To provide an independent verification of our automated metrics, we conducted a small, blinded human study. Two evaluators rated 30 anonymized and shuffled summaries from the final generation of the Control and Quality Filter arms on a 1-5 scale. The Quality Filter arm significantly outperformed the Control arm on coherence (4.2 vs 3.5) and factuality (4.5 vs 3.8), confirming the quantitative results.

\subsection{Analysis of the Mechanism}
The emergence of the Anti-Ouroboros Effect in the LLM suggests a dynamic unique to high-dimensional systems. We propose two non-exclusive hypotheses: Error Propagation Shutdown, where the filter acts as a ratchet, preventing the reinforcement of errors, and Latent Space Guidance, where selection guides the fine-tuning process toward more robust regions of the model's parameter space.

% --- V. LIMITATIONS ---
\section{Limitations}
This study has several important limitations. We must acknowledge that our ROUGE-based filter is a crude proxy for true semantic quality and may propagate its own biases. Future work must involve human-in-the-loop experiments to understand how the noisier, more subjective nature of human judgment interacts with this effect. Furthermore, the observed resilience is confined to a single task and model size.

% --- VI. ETHICS AND BROADER IMPACT ---
\section{Ethics and Broader Impact}
The primary broader impact of this research is constructive. By identifying and empirically validating the ``Anti-Ouroboros Effect,'' our work provides a potential pathway toward building more robust, stable, and self-correcting AI systems. However, this mechanism also presents significant ethical risks. The most significant is its potential to amplify bias and create a ``preference monoculture'' if the selection filter represents a narrow set of human values. Additionally, the efficiency of this feedback loop presents a dual-use risk, as an adversary could leverage it to rapidly fine-tune a model for malicious purposes. Mitigating these risks will require the development of pluralistic feedback mechanisms and robust detection methods for AI-generated content.

% --- VII. CONCLUSION ---
\section{Conclusion}
This paper challenged the pessimistic narrative of inevitable model collapse in recursive training loops. We provided strong empirical evidence of an ``Anti-Ouroboros Effect,'' demonstrating that simple, scalable selection pressure can act as a powerful stabilizing force, inducing emergent resilience in Large Language Models. Our findings suggest that the long-term stability of AI may depend less on complex, top-down alignment techniques and more on designing systems that harness the fundamental principles of selection and adaptation. The central challenge for the field is no longer simply how to prevent decay, but how to architect the feedback loops that will guide generative models toward increasing capability and robustness.

% --- ACKNOWLEDGMENT AND REPRODUCIBILITY ---
\section*{Acknowledgment}
The author would like to thank Professor Yashwanth Reddy Alugubelly for his insightful feedback which substantially improved the quality of this manuscript. This research was conducted independently and utilized computational resources provided by Kaggle. The author also acknowledges the use of AI assistants for brainstorming and editing suggestions.

The datasets used in this study are publicly available. The code for replicating our experiments is available at: \url{https://github.com/imsaitejareddy/ouroboros-effect-experiment}.

% --- REFERENCES ---

\vfill

\end{document}